\documentclass[10pt,twocolumn,letterpaper]{article}

\usepackage{cvpr}              

\usepackage{graphicx}
\usepackage{amsmath}
\usepackage{amssymb}
\usepackage{booktabs}
\usepackage{algorithm}
\usepackage{algpseudocode}
\usepackage{placeins}
\usepackage{comment}

\usepackage[pagebackref,breaklinks,colorlinks]{hyperref}

\usepackage[capitalize]{cleveref}
\crefname{section}{Sec.}{Secs.}
\Crefname{section}{Section}{Sections}
\Crefname{table}{Table}{Tables}
\crefname{table}{Tab.}{Tabs.}

\def\cvprPaperID{*****} 
\def\confName{CVPR}
\def\confYear{2023}

\begin{document}

\title{Stable Remaster: Bridging the Gap Between Old Content and New Displays}

\author{
Nathan Paull\\
{\tt\small napaull@utexas.edu}
\and
Shuvam Keshari\\
{\tt\small skeshari@utexas.edu}
\and
Yian Wong\\
{\tt\small yian@utexas.edu}
}
\maketitle

\begin{abstract}
The invention of modern displays has enhanced the viewer experience for any kind of content: ranging from sports to movies in 8K high-definition resolution. However, older content developed for CRT or early Plasma screen TVs has become outdated quickly and no longer meets current aspect ratio and resolution standards. In this paper, we explore whether we can solve this problem with the use of diffusion models to adapt old content to meet contemporary expectations.  We explore the ability to combine multiple independent computer vision tasks to attempt to solve the problem of expanding aspect ratios of old animated content such that the new content would be indistinguishable from the source material to a brand-new viewer. These existing capabilities include Stable Diffusion, Content-Aware Scene Detection, Object Detection, and Key Point Matching. We were able to successfully chain these tasks together in a way that generated reasonable outputs, however, future work needs to be done to improve and expand the application to non-animated content as well.
\end{abstract}

\section{Introduction}
\label{sec:introduction}
The way we perceive content has been revolutionized by the rapid progress of modern displays. From stunning 4K nature documentaries to sports broadcasts that provide such clarity that anyone can act as a referee, modern displays have enhanced the viewing experience. However, older content developed for CRT or early Plasma screen TVs has become outdated quickly and no longer meets current aspect ratio and resolution standards, resulting in a less enjoyable re-watching experience of beloved shows. Fortunately, we can solve this problem with the use of diffusion models to adapt old content to meet contemporary expectations.

Although Stable Diffusion\cite{https://doi.org/10.48550/arxiv.2112.10752} has gained popularity for image generation, its application to video often faces a challenge of temporal continuity. This poses a significant issue for aspect ratio expansion as the expanded content typically comprises static backgrounds. Our proposed project aims to address this limitation by utilizing the static spatial bias to govern the generation of new content and ensure that multiple instances of the same background region are not produced. By doing so, we can overcome the issue of temporal continuity in video generated using Stable Diffusion, thereby enhancing the quality and coherence of the output.

To achieve this goal, we will utilize a novel approach that combines Stable Diffusion with machine learning techniques. Specifically, we will use a machine learning model to identify and extract the static background regions from the input video. These regions will then be used as a reference to generate new content that preserves the temporal coherence of the video. Additionally, we will explore the use of other techniques such as motion estimation to further improve the quality of the output.

\section{Related Work}
\label{sec:related-work}

\subsection{Modernizing Video Techniques}
In recent years, there has been a growing interest in modernizing video through techniques such as super-resolution \cite{liu2011bayesian, Shi_2016_CVPR, kappeler2016video}, colorization \cite{yatziv2006fast, zhang2019deep}, and changing aspect ratio via outpainting \cite{guo2008aspect, soe2022content}.

Super-resolution techniques aim to increase the resolution of videos beyond their original quality. Liu et al. \cite{liu2011bayesian} proposed a Bayesian approach to adaptive video super-resolution, which estimates motion, blur kernel, and noise level while reconstructing high-resolution frames. This approach achieved promising results that can adapt to various conditions. Shi et al. \cite{Shi_2016_CVPR} used an efficient sub-pixel convolution layer for real-time super-resolution of 1080p videos on a single GPU, improving performance and reducing computational complexity compared to previous CNN-based methods. Kappeler et al. \cite{kappeler2016video} proposed a CNN for video super-resolution that combines both spatial and temporal information, achieving state-of-the-art results with a relatively small video database for training.

Colorization is the task of coloring a grayscale video. Yatziv et al. \cite{yatziv2006fast} proposed a computationally efficient method for colorizing grayscale images and videos using luminance-weighted chrominance blending and fast intrinsic distance computations, resulting in high-quality outputs with reduced computational cost and user interaction. Zhang et al. \cite{zhang2019deep} introduced an end-to-end network for video colorization that addresses the challenge of achieving temporal consistency while remaining faithful to the reference style. Their approach uses a recurrent framework that unifies semantic correspondence and color propagation steps, producing superior results compared to state-of-the-art methods.

Aspect ratio conversion is an evolving task, where older videos are adapted to fit on more modern devices with different aspect ratios. Guo et al. \cite{guo2008aspect} proposed a method for converting video aspect ratios using a saliency model to determine regions of interest and applying a novel cropping and expanding mode to maintain visual quality and avoid distortion. Soe et al. \cite{soe2022content} presented an idiom-based tool for video retargeting that allows users to control cropping and panning with selected cinematic idioms to achieve an optimal viewing experience on different platforms. However, these methods focus on cropping rather than generating new regions of the video to account for the aspect ratio change, and to our knowledge, this is one of the first works to use generative machine learning for this task.

\subsection{Background Collapsing and Stitching}

Background collapsing and stitching techniques are essential in image and video processing for tasks such as background removal, scene extension, panorama creation, and video retargeting. These techniques provide visually consistent and seamless results while maintaining the integrity of the foreground objects.

\subsubsection{Background Collapsing}
Background collapsing involves identifying and reducing redundant background regions in images or videos, allowing for the preservation of important foreground elements while resizing or retargeting. This technique often employs saliency maps or object detection algorithms to determine the importance of different regions in an image or video frame.

One prominent method for background collapsing is seam carving \cite{avidan2007seam}, which involves removing or inserting pixels along optimal seams to resize images while maintaining the essential content. Seam carving has been further extended to videos by Rubinstein et al. \cite{rubinstein2008improved}, who introduced a method for video retargeting that reduces or expands background regions while preserving the overall content and temporal coherence.

Another approach for background collapsing is patch-based image quilting \cite{efros2001image}, which synthesizes textures by sampling patches from the input image and stitching them together in a visually consistent manner. This method has been extended to videos by Kwatra et al. \cite{kwatra2003graphcut}, who introduced an algorithm for video texture synthesis using a graph-based approach to synthesize temporally coherent video textures by stitching together small spatiotemporal patches from the input video.

\subsubsection{Background Stitching}
Background stitching techniques are used to combine parts of images or video frames to create a seamless and visually consistent output. These methods are essential for tasks such as panorama creation, video compositing, and background extension.

One common approach for background stitching in images is feature-based alignment \cite{brown2007automatic}, which matches key points between overlapping regions of images and computes the transformation matrix to align and stitch the images together. This method has been further extended to videos for creating panoramic video sequences by Szeliski \cite{szeliski2006image}.

Another technique for background stitching in videos is the content-aware video retargeting method proposed by Wang et al. \cite{wang2008optimized}. This method employs a patch-based optimization approach to generate an output video with the desired target aspect ratio by stitching together patches from the input video while maintaining foreground object proportions and minimizing distortion.

Background collapsing and stitching techniques play a crucial role in image and video processing tasks. These techniques allow for the adaptation and enhancement of visual content while preserving the integrity of foreground objects and maintaining overall visual consistency. Advances in these techniques continue to improve the quality and versatility of image and video content for display on various devices and platforms.

\subsection{Stable Diffusion and Video Generation}
Image synthesis is a rapidly evolving field in computer vision, but it also has significant computational demands. Diffusion models (DMs) achieve state-of-the-art synthesis results on image data and beyond by decomposing the image formation process into a sequential application of denoising autoencoders. Robin et al. \cite{https://doi.org/10.48550/arxiv.2112.10752} presented latent diffusion models, which significantly improve the training and sampling efficiency of denoising diffusion models without degrading their quality.

Dehan et al. \cite{9857385} presented a method for video outpainting by converting a portrait (9:16) to landscape (16:9) video using background estimation, segmentation, inpainting, optical flow for temporal consistency, and image shifting to improve individual frame completions. They evaluated their method on the DAVIS and YouTube-VOS datasets. Ho et al. \cite{ho2022imagen} introduced a method to generate high-definition videos using a base video generation model and a sequence of interleaved spatial and temporal video super-resolution models. Jin et al. \cite{10031090} proposed a framework that can retarget the old video screen ratio to a wider target aspect ratio horizontally while preserving the quality of the objects using segmentation and inpainting networks. Relocating objects, especially those at the edges of the frames in the video, can be challenging.

These recent advancements in modernizing video techniques, stable diffusion, and video generation show promise in improving the visual quality and compatibility of older videos for display on modern devices. 
\section{Methods}
\label{sec:methods}

\subsection{Dataset}
Due to the nature of our project being the orchestration of several computer vision capabilities, we did not require labeled data, instead, we needed an animated video with an old aspect ratio. For our primary dataset, we have gathered videos from the animated television show 'Avatar the Last Airbender'. We have chosen this show as it has an aspect ratio of 4:3 and is popular enough to be recognized by many individuals. The image below shows how a single frame in the 4:3 aspect ratio would be displayed on many modern devices. These large black bars on the sides of the content obviously detract from the user's watching experience. 

\FloatBarrier
\begin{figure}[h]
\centering
\includegraphics[width=0.4\textwidth]{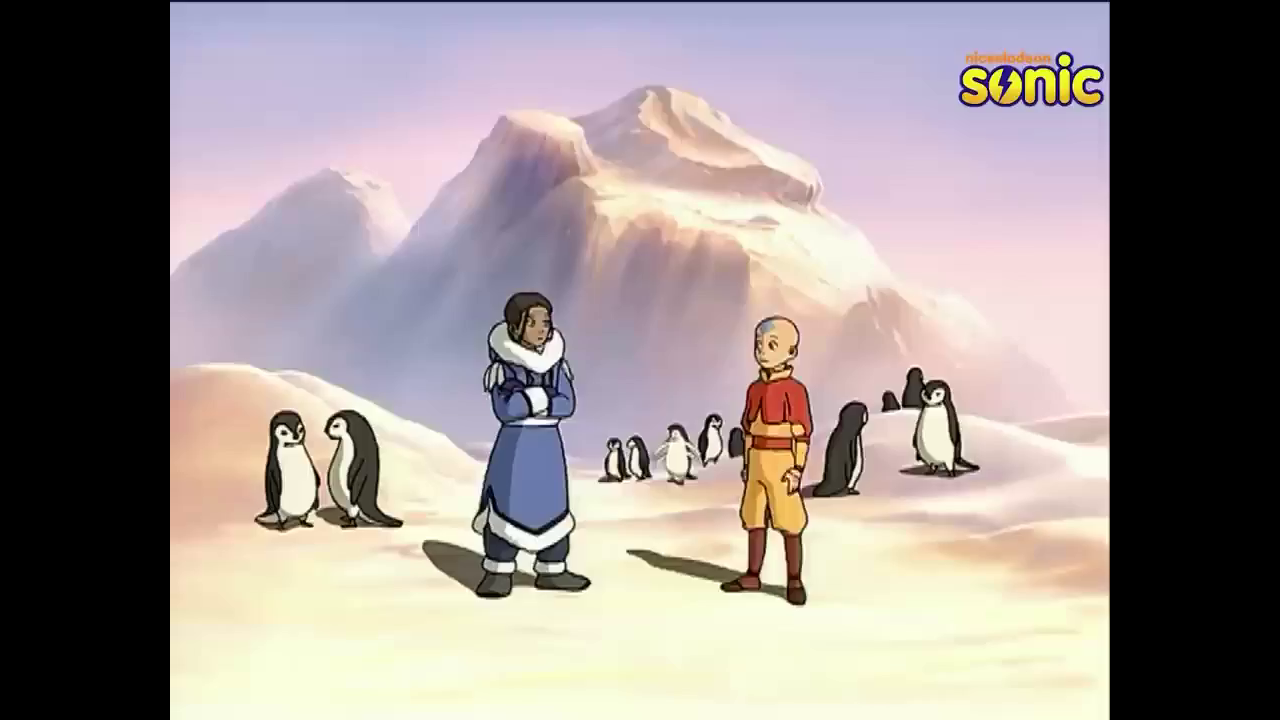}
\caption{Sample frame in 4:3 aspect ratio showing large sidebars.}
\label{fig:sample}
\end{figure}
\FloatBarrier

These large black bars are the areas we seek to fill in with content that does not distract from the existing frame but instead creates a more immersive experience. We plan to verify the accuracy/quality of the final video file generated by manual inspection as our goal is to assess human experience and immersion. As such if anything seems out of place or violates environmental rules set by the animator it would be deemed incorrect.

\subsection{Overview}
For this project, we have created a pipeline for expanding the aspect ratio of a given video which we have divided into 5 tasks which are shown in Figure 2.
\FloatBarrier
\begin{figure}[h]
\centering
\includegraphics[scale=0.98]{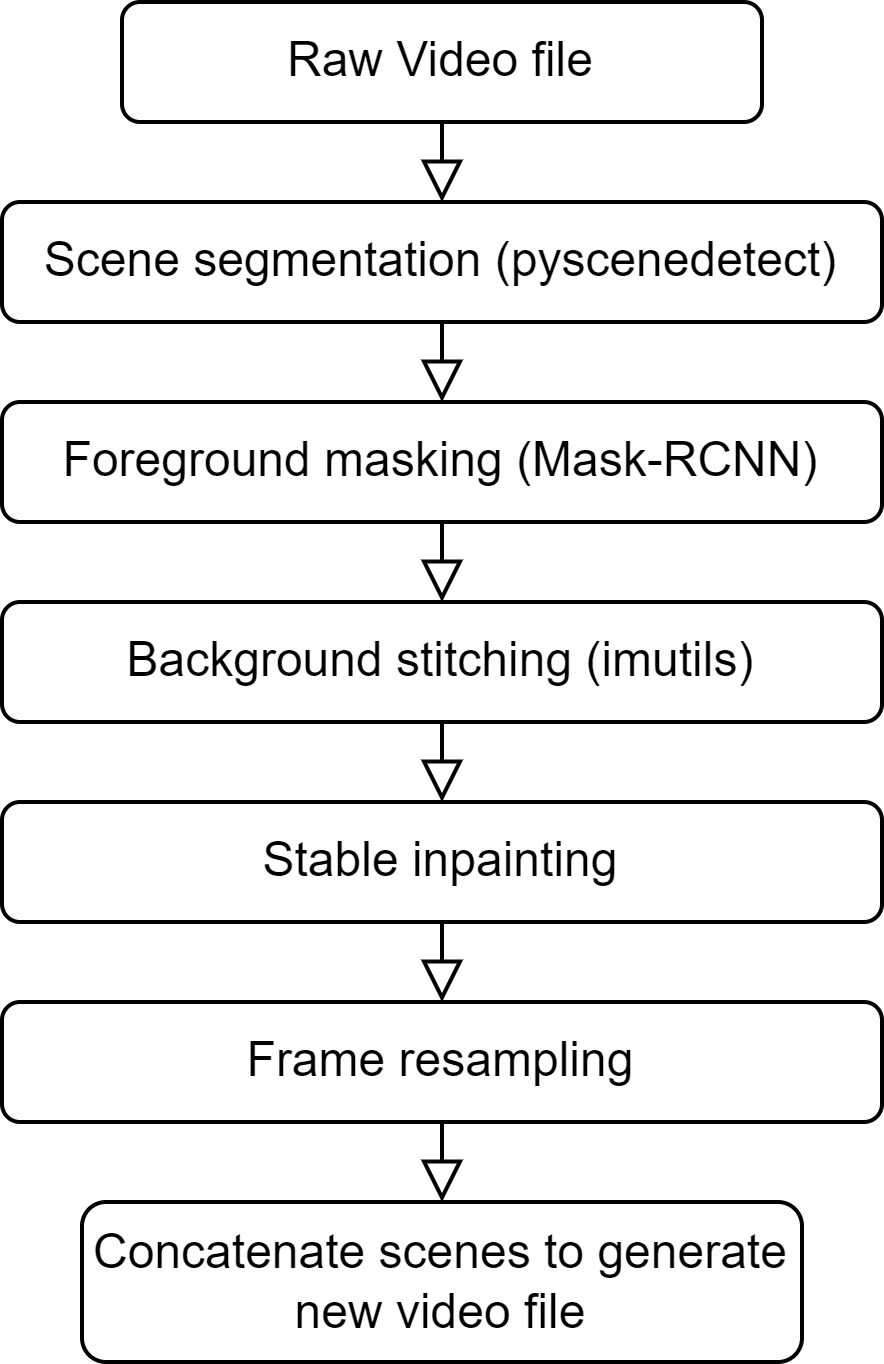}
\caption{Flowchart for the process showing the used Python libraries in brackets}
\label{fig:flowchart}
\end{figure}
\FloatBarrier
As we can see in the image the pipeline stages are Scene Segmentation, Foreground Masking, Background Stitching, Background Outpainting, and Frame Resampling. We have further broken down Background Outpainting into two sections below (Outpaint Region Selection and Background Outpainting) in spite of them being treated as one task in our code. This is in part to discuss our methods in comparison to related work and due to the fact that these two tasks are semantically different even if it doesn't make sense to separate them in the pipeline itself. 

We felt that the tasks in the pipeline were key tasks due to the following assumptions about animated content: Background within a scene is constant and maintains object permanence, camera motions generally follow affine transformations, and content at the edges of a frame will be background content. These assumptions allow us to streamline computation of this task, letting us process scenes independently, only try to generate background pixels, and stitch backgrounds together simply. There are likely some flaws in these assumptions or a set of assumptions that should be included with these above which we will discuss below in the results section as we identify shortcomings with our methodology. 

\subsection{Scene Identification and Segmentation}
For scene identification and segmentation, we chose to use PySceneDetect's Python API as it could perform the scene segmentation and save output scenes to mp4 rather than just returning frame indexes within the parent mp4 file. We felt that this was key in any future work around parallelization as all tasks after this one in the pipeline can be run on scenes in parallel, sharply decreasing the overall runtime for expanding an episode. 

When PySceneDetect is compared to alternatives such as SceneCutExtractor, and MatLabSceneDetection, and writing our own scene detection using functions within the OpenCV python API, our selection was quickly narrowed down to PySceneDetect and SceneCutExtractor. First, we felt the Python API offered by both of these libraries was crucial to making the pipeline easy to use and edit. Additionally, the ability to use a prepackaged library gave us much more time to work on the pipeline itself rather than focusing on just a single task. In our research, PySceneDetect was preferable to SceneCutExtractor because of its ability to save scenes to mp4, while SceneCutExtractor would save JSON or CSV files with frame indexes and evaluations. While this may provide more flexibility, we found the ease of use for PySceneDetect to be much more attractive. 

Within PySceneDetect we used the content-aware scene detection which detects changes in the HSV color space to determine scene changes. Additionally, PySceneDetect uses ffmpeg to perform the scene cuts.

\subsection{Foreground Masking}
In this section of the pipeline, we make one further assumption, that objects in the foreground will first be rendered in full by the animator in addition the the assumption that objects in the background are permanent throughout the duration of a scene. This assumption motivates the idea of foreground masking so that we solely analyze the background when generating pixels. 

To perform this masking we sought out a network that could recognize objects in the foreground, find bounding boxes for these objects (or masks if possible), and perform these two tasks at a relatively quick speed. We decided that bounding boxes was a minimum requirement as anything less than that would not allow us to run an algorithm such as GrabCut, however, if the method was able to generate masks on its own the need for GrabCut would be removed. These requirements led us to select Mask-RCNN, an object detection DNN built on a base ResNet Structure. Mask-RCNN not only detects a vast array of objects found in the COCO dataset but additionally generates masks for these objects and can be run at a speed of at least 5 frames per second on most GPUs. In our deployment on an Nvidia 2070 Super, we were able to get a speed of 7 frames per second. 

With the selection of Mask-RCNN, all we had to do was to combine the masks of found foreground objects into a total mask that would separate the background from the foreground for the next stage in the pipeline, background stitching. 

\subsection{Background Stitching}
The goal of background stitching was to create a total background for the scene. This is motivated by a common technique in the film used for creating semi-transparent characters. To accomplish this, two shots will be filmed sequentially, one with the character in the frame and one without the character in the frame. This allows editors to fill in the missing information with real information instead of generated information. Our goal was similar, to avoid using generated pixels wherever possible. This is because generating coherent pixels is computationally expensive and because we want to maintain pixels that exist in the original animation. If we generate the legs of a table on the boundary of a scene only for the camera to pan towards that table and have it disappear behind the original frame, we would immediately break immersion for the viewer. 
\FloatBarrier
\begin{figure}[h]
\centering
\includegraphics[width=0.98\linewidth]{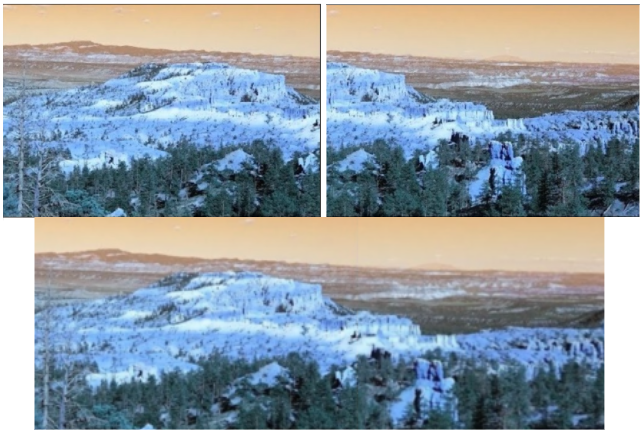}
\caption{Keypoint matching to extend the aspect ratio (generating a panorama: bottom) from two slightly different images as can be expected from consecutive frames (top 2 images) in a scene}
\label{fig:stitch}
\end{figure}
\FloatBarrier
Through this motivation for background stitching, we decided to use keypoint matching and affine transformation estimation. This method of keypoint matching and transformation estimation is common when generating panoramic photos from a set of distinct images. As such we chose to follow this method choosing SIFT as our method for keypoint detection and description. With these SIFT key points, we could then find a set of good matches to determine any affine transformations necessary for matching the set of images together.

Once we achieved a complete background we could then accurately determine which pixels can be sampled from information generated by the original animators and which pixels would need to be generated. 

\subsection{Outpaint Region Selection}
Alongside the total background that we have constructed, we similarly construct a total mask. This mask will contain information regarding which parts of the background have been filled with information derived from frames within the scene and which parts of the background lack any information. This total mask allows us to determine which regions of the total background will require generated pixels.

Pixels will only be generated if they fall within the bounds of the new frames, this means that each time we will sample this total mask within the region of the new frame to determine if new pixels need to be generated. If no new pixels are needed then the answer is to simply sample the total background and return. If new pixels are needed however we will use this sampled mask to inform any outpainting and then we will add these generated pixels to the total background and update the total mask so that these pixels will not be generated again. In doing this we decrease the number of pixels that must be generated.

\subsection{Background Outpainting}
For the task of outpainting, we will use the practice of Stable Diffusion\cite{https://doi.org/10.48550/arxiv.2112.10752} as implemented by the diffusers python library\cite{von-platen-etal-2022-diffusers}. This library provides pre-trained models that can be used through a simple Python API. Specifically, we chose the Stable Diffusion Inpainting Pipeline offered by this library as it allows the user to input a mask where pixels should be generated and allows for the user to input a prompt that will describe the generated pixels. While optimization of the prompt would likely improve results, we decided to use the generic prompt of 'animated background' for all generated pixels in the hope that it would create reasonable generated pixels. These generated pixels are then added to the total background generated from background stitching which will adjust the Outpaint Region selected for the next frame within the same scene.

This step of the pipeline we found to be our most time-consuming taking up to 40 seconds per frame to generate pixels. This in part is why some of the previous steps are necessary. Without any of the previous steps, a 20-minute episode at 30 frames per second would take nearly 400 hours to process. While we cannot provide a tighter upper bound than this, the lower bound is much lower as there is no generation of duplicate pixels. This means that longer scenes create shorter runtimes as no duplicate pixels are generated. This could likely be further optimized by choosing the frames that experience large translation transformations relative to the first frame along a set of key directions. 

\subsection{Frame Resampling}
After we have generated all necessary pixels to fill in gaps within the background the task of frame resampling is rather simple. We begin by using the  affine transformation found in the background stitching step to transform our sampling region. We then simply select all pixels in the total background that are within this region. Following this we calculate the inverse of this affine transformation and use this inverse to transform our sampled pixels back into a viewable frame. We continue this for each frame until the scene has been completely reconstructed. We can now save these frames in a new scene and then concatenate all scenes together to create the reconstructed episode. 

\subsection{Experiments and Results}

Our primary stated goal was to create a pipeline that could expand the aspect ratio of old animated content without violating the temporal coherency of the content. We believe that overall we were successful in this endeavor with the results shown below.

\FloatBarrier
\begin{figure}[h]
\centering
\includegraphics[width=0.98\linewidth]{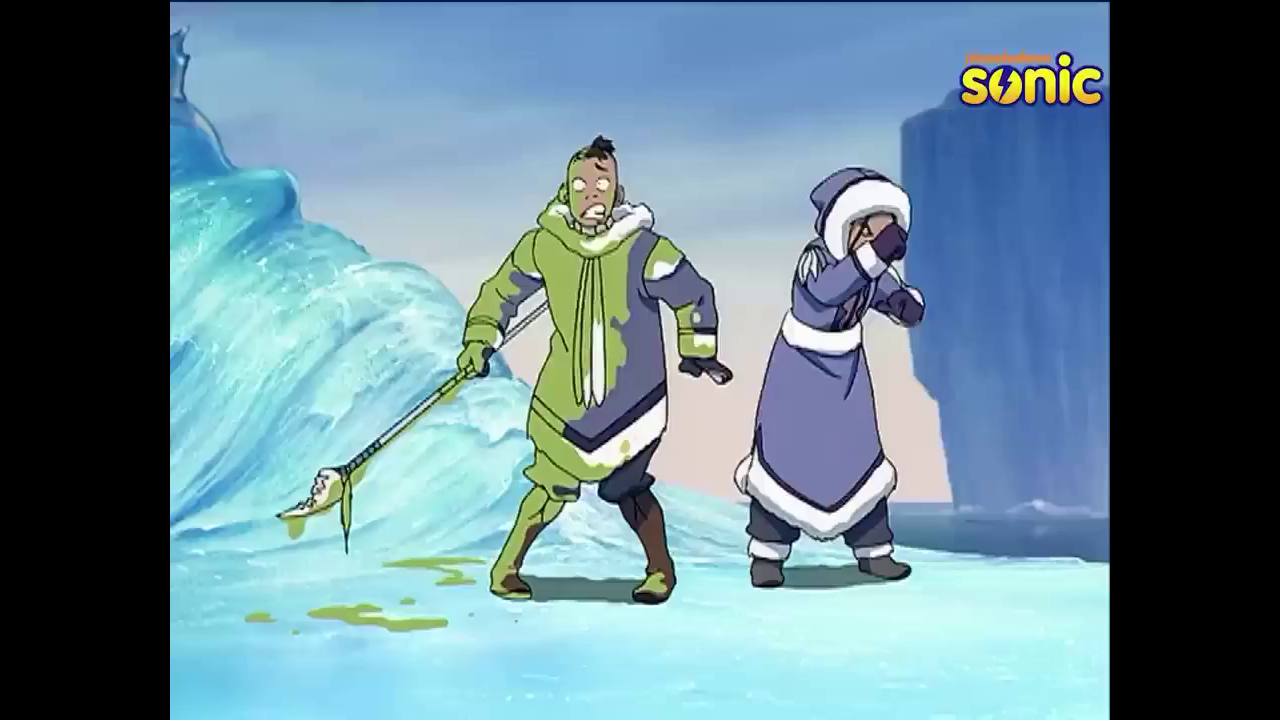}
\includegraphics[width=0.98\linewidth]{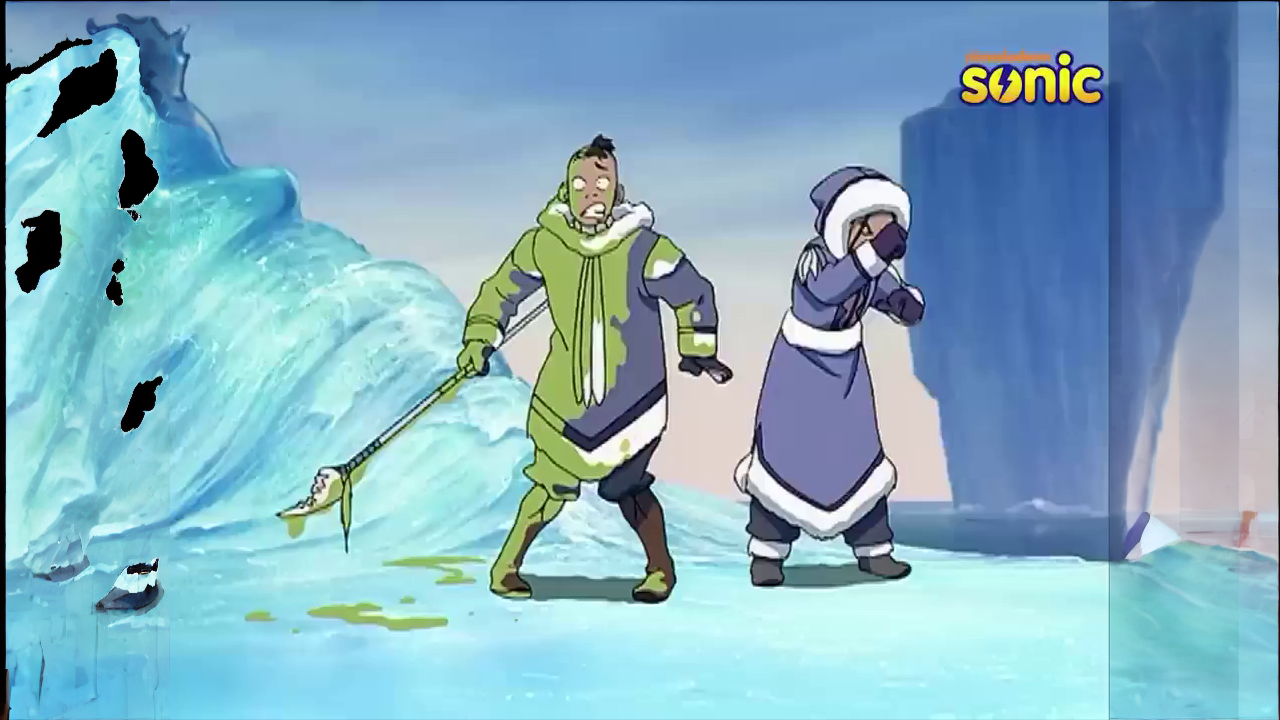}
\caption{Comparison of the original frame (top) with resampled frame (bottom)}
\label{fig:iceberg}
\end{figure}
\FloatBarrier
\FloatBarrier
\begin{figure}[h]
\centering
\includegraphics[width=0.98\linewidth]{images/padded_snow.png}
\includegraphics[width=0.98\linewidth]{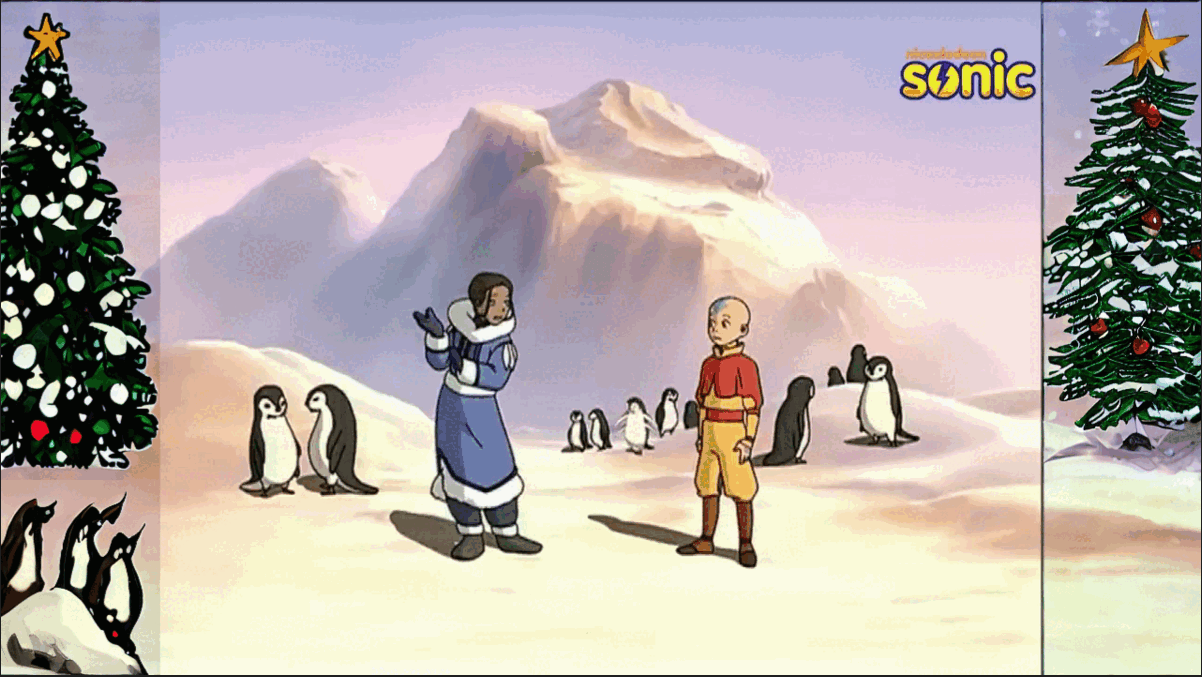}
\caption{Comparison of the original frame (top) with resampled frame (bottom)}
\label{fig:snow}
\end{figure}
\FloatBarrier
In both of the figures above we can see some similar results. The easiest to see is the shortcomings with Figure 4 showing black spots on the left side of the frame and Figure 5 adding Christmas trees and distorted penguins. Additionally, both images seem to have some color distortion and both possess a vertical gray line on the right-hand side of the image. We also must admit that the color distortion is not constant throughout the scene either and this can be seen in the gif results linked in the Demo section below. All of these shortcomings do bring us short of the goal of perfectly adapting old animated content for modern screens with modern aspect ratios. 

However, only mentioning these shortcomings would be ignoring the parts of the expansion that the pipeline gets correct. If you specifically look at the floor in both figures you can see how the scene is properly expanded. In the second figure, the mounds of snow are continued to the boundaries of the frame quite well and show that while this method was not completely successful that it is quite close. In the Future Work section below we continue to discuss what we believe can be done to solve these shortcomings and solve the stated goal. 

While we can see the effects of Stable Diffusion on the figures above, we would like to specifically discuss some results from individual steps earlier in the pipeline so that we can analyze which steps are creating the error we see in the final product. We will first discuss the Foreground Masking section of the pipeline.
\FloatBarrier
\begin{figure}[h]
\centering
\includegraphics[width=0.98\linewidth]{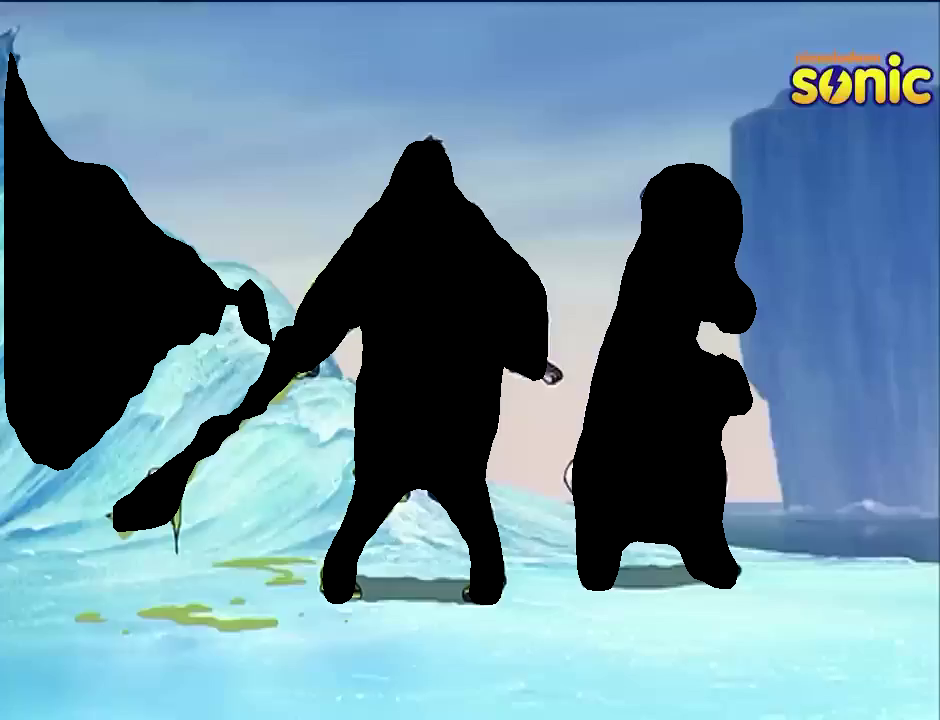}
\caption{False objects masked}
\label{fig: false mask}
\end{figure}
\FloatBarrier

As we can see in the figure above, Mask-RCNN seems to detect some false objects in the animated domain. Additionally, we can see in the figure below how Mask-RCNN can also fail to detect objects when animators break body continuity to better display character motion.

\FloatBarrier
\begin{figure}[h]
\centering
\includegraphics[width=0.98\linewidth]{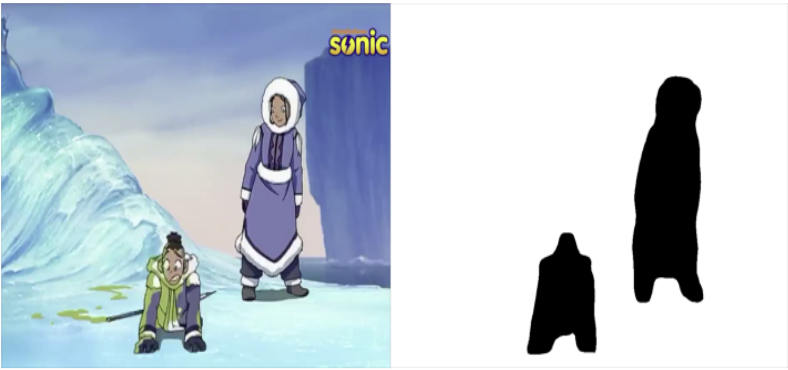}
\caption{Foreground masking}
\label{fig:mask}
\end{figure}
\FloatBarrier
Lastly, we can see how some of the object boundaries are very slightly off. While our method for background concatenation can help to limit some of this noise if several frames occur at similar camera positions, it is still possible for this noise to persist and affect certain frames.Overall these three results show that Mask-RCNN does not perform perfectly and adjustment or replacement of this model could lead to improved results at the output of the pipeline. 

Finally, we would like to cover the results of the Background Stitching module. This module is what we believe to be the source of most of the shortcomings, especially the shortcomings related to color change around the edges of the image. 

\FloatBarrier
\begin{figure}[h]
\centering
\includegraphics[width=0.98\linewidth]{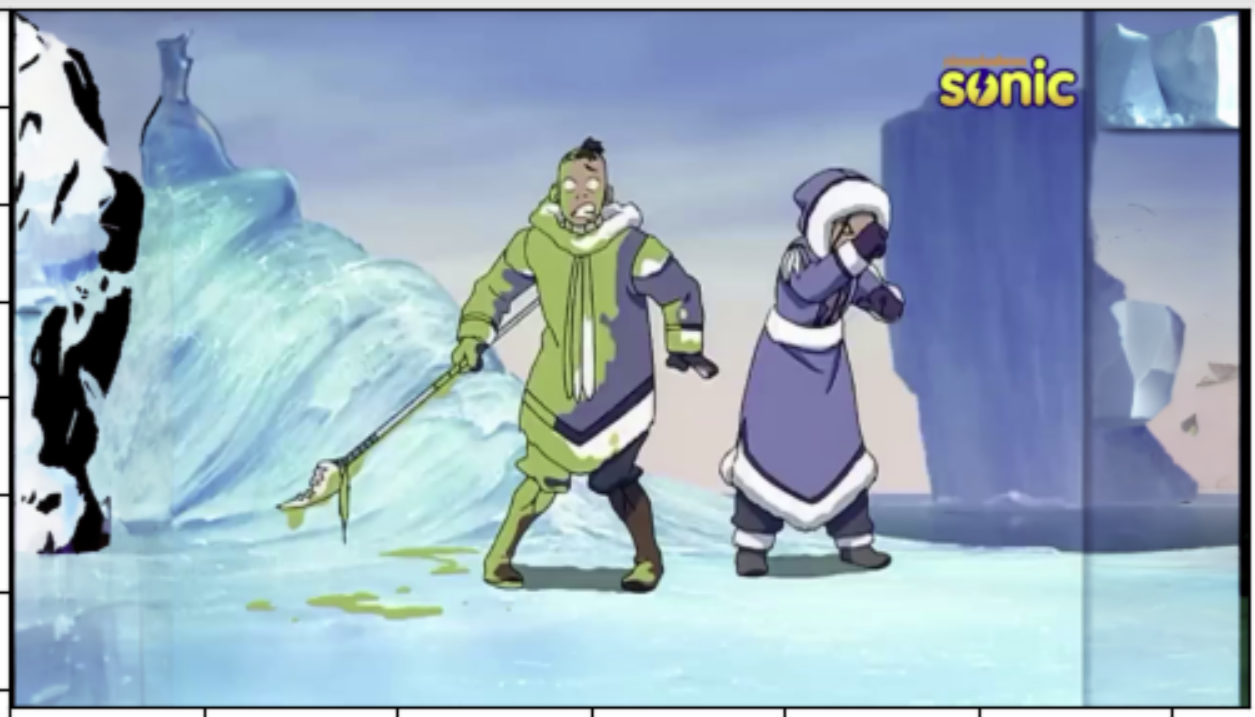}
\caption{Sample Frame displaying edge effects on final output.}
\label{fig:iceberg2}
\end{figure}
\FloatBarrier
We believe that these off-color effects are due to the affine transformations performed in this step. In addition to these color effects, we also experience issues with key point matching on repetitive backgrounds where key points are harder to uniquely match. We believe that this and all other issues mentioned above are what lead to most lackluster results, but we are still excited by what all was in fact accomplished by this pipeline.

\subsection{Demo}
Our code is available as a GitHub repository named {\href{https://github.com/naston/StableRemaster}{Stable Remaster}}. This Github provides instructions on how to set up the environment and run the demonstration code. The environment can be set up by creating an anaconda environment from the environment.yml file or by manually installing the libraries listed in that file. There is also the need to install ffmpeg for the full pipeline demonstration but this is not necessary for the scene-based demonstration. To run the full demonstration use the pipeline\_demo.py file. To run the scene-based demonstration use the scene\_demo.py file.

To view some example output from the scene\_demo.py demonstration please visit the following {\href{https://imgur.com/a/dPxylzY}{link}}. Here we have two gifs showing expanded scenes as well as images of the original scenes.

\section{Discussion}
\label{sec:discussion}

\subsection{Future Work}
There is a large amount of work that can be done to improve upon what we have accomplished in this paper. To start this pipeline has been designed in a modular manner so that future implementations can switch out components for more optimized versions. A primary example of this would be to remove Mask-RCNN, which performs object detection, with a model that performs the similar task of foreground detection. As discussed above Stable Diffusion composes a majority of our compute time, taking up to 40 seconds per frame. This is dramatically slower than all other stages in the pipeline and could be swapped out in favor of a faster algorithm. However, there is also room for fine-tuning of models like Mask-RCNN and Stable Diffusion for use in the animated domain. This would likely see much better results as animators often break the continuity of a characters body in order to accurately display motion. 

As far as further replacement of pipeline steps we believe that there is a need for replacement of our implementation of background stitching as it relies of SIFT key point matching which fails on repetitive backgrounds like brickwork. This specifically led to huge issues with scenes where the camera panned across a semi-repetitive background. Similarly there is work to be done to limit the noisy affects of affine transformations on the output of the pipeline. This could even include the addition of a de-noising step at the end of the pipeline to reduce the affects of the many affine transformations. Some similar tasks to this would include 

Additional extra steps in this pipeline could include tasks such as resolution scaling. We believe resolution scaling could provide two valuable capabilities. The first capability is to update resolutions to match modern displays in the same way we are updating the aspect ratio. Many displays work best with HD resolutions which many old animations do not have. Secondly, we believe down sampling could provide simpler tasks to models like Stable Diffusion, requiring the generation of fewer pixels which can then be scaled up to match a desired resolution. This would do a lot to help the computation bottleneck that we experience when running stable diffusion.

Another way to relax this computation constraint would be implementation of task parallelization. As each scene is treated independently it is not required that the computation of any two scenes be done sequentially. By running certain stages of the pipeline on scenes in parallel one could drastically cut down on the runtime of the entire pipeline. This however does make an assumption that certain scenes do not share a background, an assumption that we believe could be explored to determine which scenes would benefit from combined computation, allowing for temporal coherence across scenes and not just within them.

Lastly we believe that there is future work to be done on items that lie between the categories of background and object such as fire, lightning, and rain. These are items that should move with each frame, violating our assumption of a static background. Another assumption that is violated is the assumption of affine camera transformations. While neither of these assumptions are violated often, it is a valuable area of study to make this pipeline more robust.

\subsection{Conclusion}
In this paper we explore the ability to combine multiple independent computer vision tasks to attempt to solve the problem of expanding aspect ratios of old animated content such that the new content would be indistinguishable from the source material to a brand new viewer. These existing capabilities include Stable Diffusion, Content Aware Scene Detection, Object Detection, and Key Point Matching. 
While we did successfully chain these tasks together in a way that generated reasonable output, we were not successful in this task. However, we still feel that the pipeline we have constructed pipeline serves as a great foundation for future work. Allowing for the introduction of new stages in the process or the replacement of old stages.

{\small
\bibliographystyle{ieee_fullname}
\bibliography{refs.bib}

\begin{thebibliography}{10}\itemsep=-1pt

\bibitem{avidan2007seam}
Shai Avidan and Ariel Shamir.
\newblock Seam carving for content-aware image resizing.
\newblock {\em ACM Transactions on Graphics (TOG)}, 26(3):10--es, 2007.

\bibitem{brown2007automatic}
Matthew Brown and David~G. Lowe.
\newblock Automatic panoramic image stitching using invariant features.
\newblock In {\em International Journal of Computer Vision}, volume~74, pages
  59--73. Springer, 2007.

\bibitem{9857385}
Loïc Dehan, Wiebe Van~Ranst, Patrick Vandewalle, and Toon Goedemé.
\newblock Complete and temporally consistent video outpainting.
\newblock In {\em 2022 IEEE/CVF Conference on Computer Vision and Pattern
  Recognition Workshops (CVPRW)}, pages 686--694, 2022.

\bibitem{efros2001image}
Alexei~A Efros and William~T Freeman.
\newblock Image quilting for texture synthesis and transfer.
\newblock In {\em Proceedings of the 28th annual conference on Computer
  graphics and interactive techniques}, pages 341--346. ACM, 2001.

\bibitem{guo2008aspect}
Yandong Guo, Zhongliang Deng, Xiaodong Gu, Zhibo Chen, Quqing Chen, and Charles
  Wang.
\newblock Aspect ratio conversion based on saliency model.
\newblock In {\em 2008 Congress on Image and Signal Processing}, volume~4,
  pages 92--96. IEEE, 2008.

\bibitem{ho2022imagen}
Jonathan Ho, William Chan, Chitwan Saharia, Jay Whang, Ruiqi Gao, Alexey
  Gritsenko, Diederik~P Kingma, Ben Poole, Mohammad Norouzi, David~J Fleet,
  et~al.
\newblock Imagen video: High definition video generation with diffusion models.
\newblock {\em arXiv preprint arXiv:2210.02303}, 2022.

\bibitem{10031090}
Jun-Gyu Jin, Jaehyun Bae, Han-Gyul Baek, and Sang-Hyo Park.
\newblock Object-ratio-preserving video retargeting framework based on
  segmentation and inpainting.
\newblock In {\em 2023 IEEE/CVF Winter Conference on Applications of Computer
  Vision Workshops (WACVW)}, pages 497--503, 2023.

\bibitem{kappeler2016video}
Armin Kappeler, Seunghwan Yoo, Qiqin Dai, and Aggelos~K Katsaggelos.
\newblock Video super-resolution with convolutional neural networks.
\newblock {\em IEEE transactions on computational imaging}, 2(2):109--122,
  2016.

\bibitem{kwatra2003graphcut}
Vivek Kwatra, Arno Sch{\"o}dl, Irfan Essa, Greg Turk, and Aaron Bobick.
\newblock Graphcut textures: image and video synthesis using graph cuts.
\newblock In {\em ACM Transactions on Graphics (TOG)}, volume~22, pages
  277--286. ACM, 2003.

\bibitem{liu2011bayesian}
Ce Liu and Deqing Sun.
\newblock A bayesian approach to adaptive video super resolution.
\newblock In {\em CVPR 2011}, pages 209--216. IEEE, 2011.

\bibitem{https://doi.org/10.48550/arxiv.2112.10752}
Robin Rombach, Andreas Blattmann, Dominik Lorenz, Patrick Esser, and Björn
  Ommer.
\newblock High-resolution image synthesis with latent diffusion models, 2021.

\bibitem{rubinstein2008improved}
Michael Rubinstein, Ariel Shamir, and Shai Avidan.
\newblock Improved seam carving for video retargeting.
\newblock In {\em ACM Transactions on Graphics (TOG)}, volume~27, pages 1--9.
  ACM, 2008.

\bibitem{Shi_2016_CVPR}
Wenzhe Shi, Jose Caballero, Ferenc Huszar, Johannes Totz, Andrew~P. Aitken, Rob
  Bishop, Daniel Rueckert, and Zehan Wang.
\newblock Real-time single image and video super-resolution using an efficient
  sub-pixel convolutional neural network.
\newblock In {\em Proceedings of the IEEE Conference on Computer Vision and
  Pattern Recognition (CVPR)}, June 2016.

\bibitem{soe2022content}
Than~Htut Soe and Marija Slavkovik.
\newblock A content-aware tool for converting videos to narrower aspect ratios.
\newblock In {\em ACM International Conference on Interactive Media
  Experiences}, pages 109--120, 2022.

\bibitem{szeliski2006image}
Richard Szeliski.
\newblock Image alignment and stitching: A tutorial.
\newblock In {\em Foundations and Trends{\textregistered} in Computer Graphics
  and Vision}, volume~2, pages 1--104. Now Publishers, 2006.

\bibitem{von-platen-etal-2022-diffusers}
Patrick von Platen, Suraj Patil, Anton Lozhkov, Pedro Cuenca, Nathan Lambert,
  Kashif Rasul, Mishig Davaadorj, and Thomas Wolf.
\newblock Diffusers: State-of-the-art diffusion models.
\newblock \url{https://github.com/huggingface/diffusers}, 2022.

\bibitem{wang2008optimized}
Yu-Shuen Wang, Chiew-Lan Tai, Olga Sorkine, and Tong-Yee Lee.
\newblock Optimized scale-and-stretch for image resizing.
\newblock In {\em ACM Transactions on Graphics (TOG)}, volume~27, pages 1--8.
  ACM, 2008.

\bibitem{yatziv2006fast}
Liron Yatziv and Guillermo Sapiro.
\newblock Fast image and video colorization using chrominance blending.
\newblock {\em IEEE transactions on image processing}, 15(5):1120--1129, 2006.

\bibitem{zhang2019deep}
Bo Zhang, Mingming He, Jing Liao, Pedro~V Sander, Lu Yuan, Amine Bermak, and
  Dong Chen.
\newblock Deep exemplar-based video colorization.
\newblock In {\em Proceedings of the IEEE/CVF conference on computer vision and
  pattern recognition}, pages 8052--8061, 2019.

\end{thebibliography}
}

\end{document}